\def\year{2021}\relax
\documentclass[letterpaper]{article} 
\usepackage{aaai21}  
\usepackage{times}  
\usepackage{helvet} 
\usepackage{courier}  
\usepackage[hyphens]{url}  
\usepackage{graphicx} 
\usepackage{graphicx}  
\usepackage{natbib}  
\usepackage{caption} 
\usepackage{todonotes}

\usepackage{color}
\usepackage{amsmath}
\usepackage{algorithm}
\usepackage{amsfonts}
\usepackage{subcaption}

\newtheorem{problem}{Problem}
\DeclareMathOperator{\EX}{\mathbb{E}}

\title{Test-Cost Sensitive Methods for Identifying Nearby Points }

\author{
    Seung Gyu Hyun,\textsuperscript{\rm 1}
    Christopher Leung\textsuperscript{\rm 2} \\
}
\affiliations{
   \textsuperscript{\rm 1} University of Waterloo \\
    \textsuperscript{\rm 2} Georgia Institute of Technology \\
    sghyun@uwaterloo.ca, cleung62@gatech.edu
}

\newcommand{\features}{\mathcal{F}}
\newcommand{\states}{\mathcal{S}}
\newcommand{\actions}{\mathcal{A}}
\newcommand{\nodes}{\mathcal{N}}
\newcommand{\rank}{\mathfrak{R}}
\newcommand{\node}{N}
\DeclareMathOperator*{\argmax}{arg\,max}
\DeclareMathOperator*{\argmin}{arg\,min}

\begin{document}

\maketitle

\begin{abstract}
Real-world applications that involve missing values are often constrained by the cost to obtain data. Test-cost sensitive, or costly feature, methods additionally 
consider the cost of acquiring features. Such methods have been extensively
studied in the problem of classification. In this paper, we study a related
problem of test-cost sensitive methods to identify nearby points from a large
set, given a new point with some unknown feature values. We present two models,
one based on a tree and another based on Deep Reinforcement Learning. In our simulations, we show that the models
outperform random agents on a set of five real-world data sets.
\end{abstract}

\section{Introduction}
In many real-world setting, resources, such as money, energy, time, 
limit the amount of information one can ascertain. 
Medical practitioners focus on arriving at a diagnosis as quickly
and cost efficiently as possible. Survey makers aim to design a small yet revealing
set of questions. Pilots require making split-second decisions by using a limited
set of measurements from a multitude of sensors, among many other examples. In
all these cases, there are two forces at play: on one hand, the agent tries to minimize
the total cost of the features it gathers, while simultaneously optimizing its ability to make
the correct decision. When applied to classification, this gives rise to the Classification
with Costly Features (CwCF) (or Test-Cost Sensitive Classification) problem, which
has been extensively studied by many researchers.

In this setting, an algorithm classifies an instance using only the features it chose to reveal at 
defined costs. For each instance, the algorithm sequentially selects the next feature to obtain
conditioned on the currently available information. This sets test-cost sensitive methods apart
from feature selection/pruning: the set of features to reveal is inherently local to the instance.
Keeping the same setting of costly features, this paper tackles a simple yet surprisingly difficult problem:
\begin{problem}
    Given a complete data set (where values for all features are known) $D$,
    an instance $p$ with some unknown feature values drawn from the same distribution, and
    a budget $B$, find the optimal set of features to reveal with total cost $\le B$ such
    that instances in $D$ closest to the completed $p$ (in distance) can be identified.
    \label{problem:main}
\end{problem}
In healthcare, this could help find similar patients for prognostic purposes,
while in recommender systems, this could assist in early detection of similar
users. This problem is also closely linked to a data repair problem of 
\emph{data imputation} since knowing the most similar and dissimilar points 
can be used to complete the missing features. Human-in-the-loop feature revealing
methods (without consideration for feature costs) 
have been successfully applied to data repair systems, such as the Guide
Data Repair (GDR) system of \cite{yakout2011guided}.

The difficulty of this problem comes from the fact we are attempting to measure distance
using a lossy embedding. For example, given two complete points $c_1 = (1,0)$, $c_2 = (2,2)$, 
and an incomplete point $p = (1,?)$, there is no way of knowing whether $p$ is closer to 
$c_1$ or $c_2$ if the distribution of the $y$-coordinate is random. However, if
$\Pr(p.y < 5/4 | x=1) > \Pr(p.y \ge 5/4|x=1)$, then it is more \emph{probable} that $p$ is
closer to $c_1$. Furthermore, if $p.y$ is functionally dependent on $p.x$, simply knowing
$p.x$ would be enough to compute the distance. In this view, Problem \ref{problem:main} can be 
restated as finding the optimal set of features such that projecting $D$ to a feature space with 
only the known features preserves the relative distances from $p$ to the points of $D$ well. 
While such optimal sets may not exist for random
distributions, we will show that this often holds for many real-life data sets.

Test-cost sensitive learning can be cast as a Markov Decision Process (MDP) 
\cite{zubek2004pruning}, and test-cost sensitive algorithms typically
fall into two categories: inductive, often tree-based, learning 
\cite{turney2002types,chai2004test,maliah2018mdp} 
and reinforcement learning \cite{dulac2011datum, janisch2019classification}. While 
\citet{janisch2019classification} shows that Deep Q-learning outperforms problem
specific inductive learning models, explainability is desired in many domains. Following these 
previous works, we present two models: a subspace clustering model based on the
CLusterTree (CLTree) model \cite{liu2005clustering} and a Deep Q-learning model. 
We evaluate these models using public data sets against random agents and show that both models 
outperform random agents. We are not aware of any work solving the same problem that we can
benchmark against.

\section{Problem formulation}
We begin by discussing how to describe a solution to Problem \ref{problem:main}. While the obvious
choice is to compute something akin to the $k$ nearest neighbours of an instance
$p$, this creates a large output space which makes learning harder. Instead,
we take a more generalizable approach and return nearby clusters after partitioning
$D$ using a distance-based clustering algorithm. Furthermore, if the notion of
membership for an arbitrary points exists (as is the case of $k$-means and
CLTree), we can return $p$'s most probable cluster. This provides a succinct output
that still allows users to retrieve similar points in $D$.

Given clusters $C_1,\dots, C_m$ of $D$ and an instance $p$ with only features
$\bar\features \subseteq \features = \{f_1,\dots, f_n \}$ known, we need to measure how well
$\bar\features$ preserves relative distances. Let $\rank_{\bar\features,i}(p)$ be
a prediction of the ranking of $C_i$ 
and $\rank_{\features,i}(p)$ be its true ranking, such that $C_i$ with rank $1$ contains
elements that are most similar to $p$ and rank $n$ most dissimilar (for some measure
of similarity).
We define the score function $S(\bar\features, p)$ as the
mean squared error (MSE) of the rankings:

\begin{align}
S(\bar\features, p) = \frac{1}{m} \sum_i (\rank_{\bar\features,i}(p) - \rank_{\features, i}(p))^2. 
\label{equation:score}
\end{align}

When $S(\bar\features,p) = 0$, it means that measuring the similarity between $p$ and
each cluster is no different if we use $\bar\features$ or all the features $\features$.
Let $c: \features \mapsto \mathbb{R}$
the cost function mapping feature $f_i$ to some real-valued normalized cost 
(that is, $1 \ge c(f) \ge 0$  for all features).
Finally, we want to find parameters $\theta$ for function $z_\theta(p) = \bar\features_{p}$
that minimizes the expected sum of MSE and (scaled) total cost:
\begin{align}
   \argmin_{\theta} 
    \frac{1}{|D|} \sum_{p_j \in D} \left( \left(\sum_{f\in\bar\features_{p_j}} \alpha c(f) \right) + 
    S(\bar\features_{p_j}, p_j) \right) 
    \label{equation:cost}
\end{align}

To formulate Problem \ref{problem:main} as a MDP, 
let $(x_1,\dots,x_n) \in D$ be a point in $D$, where $0\le x_i \le 1$ is a 
normalized value of feature $f_i \in \features$.
Let $B \in \mathbb{R}$, called the budget, be the maximum allowed cost incurred by
the agent. We define the state space as 
$$\states = \{ ((x_1,\dots,x_n), \bar \features, \bar c) \},$$ 
where $\bar\features \subseteq \features$ is the current set of selected features
and $\bar c = \sum_{f_i \in \bar \features} c(f_i)$ is the current cost.
A state $s \in \states$ is terminal if there exist no feature $f_j\notin \bar\features$
such that $\sum_{f_i \in \bar \features} c(f_i) + c(f_j) \le B$. 

The set of available actions is $\actions = \actions_f \cup \{\actions_t\}$, where
$\actions_f = \{ f_1,\cdots, f_n \}$ reveals feature $f_i$ and 
$\actions_t$ terminates the episode.
The reward function $r: \states \times \actions \mapsto \mathbb{R}$ is
$$ r(s,a) = \begin{cases}
-\alpha c(a) & \text{if $a\in \actions_f$} \\
-S(\bar\features,p) &\text{if $a = \actions_t$}
\end{cases}
$$
The parameter $\alpha$ balances the cost with the score: higher values of $\alpha$ will
make the agent favour lower cost episodes and vice versa. Starting from initial state
$s_{0,p} = (p, \phi, 0), p \in D$ (where no features are known), let
$s_{t,p} = (p, \bar\features_p, \bar c_p)$ be the terminal state following the
optimal policy $\pi_\theta$ that maximizes the sum of rewards for this episode:
$$ \sum_{a_i} r(s_i,a_i) = - \alpha \bar c_p - S(\bar\features_p,p). $$
The expected sum of rewards over all $p \in D$ is
$$ \frac{1}{|D|} \sum_p (- \alpha \bar c_p - S(\bar\features_p,p)).  $$
This is maximized when equation \ref{equation:cost} is minimized for fixed budget
$B$; 
thus, finding the optimal policy to the MDP solves Problem \ref{problem:main}.

\section{Inductive learning}
The CLTree model of \cite{liu2005clustering} uses a technique called \emph{subspace
clustering}, which was developed to effectively cluster high-dimensional data.
Subspace clustering algorithms project different partitions of the data to different
subspaces and find clusters within those selected subspaces. 
This allows the algorithm to judge if features are important \emph{locally},
conditioned on being in some subspace; thus, this is a very natural framework to solve
Problem \ref{problem:main}. For a survey of this topic, we direct the readers to \cite{parsons2004subspace}.

In this section, we present a novel subspace clustering model, 
named Cost Balancing Clustering Tree (CBCTree), which can be see as an extension of the
CLTree model that accounts for costly features. In order to facilitate this, we have to use a different 
measure of gain than the CLTree. Furthermore, we must also define cluster membership with unknown 
values. 

\begin{figure}[H]
    \centering
    \includegraphics[scale=0.25]{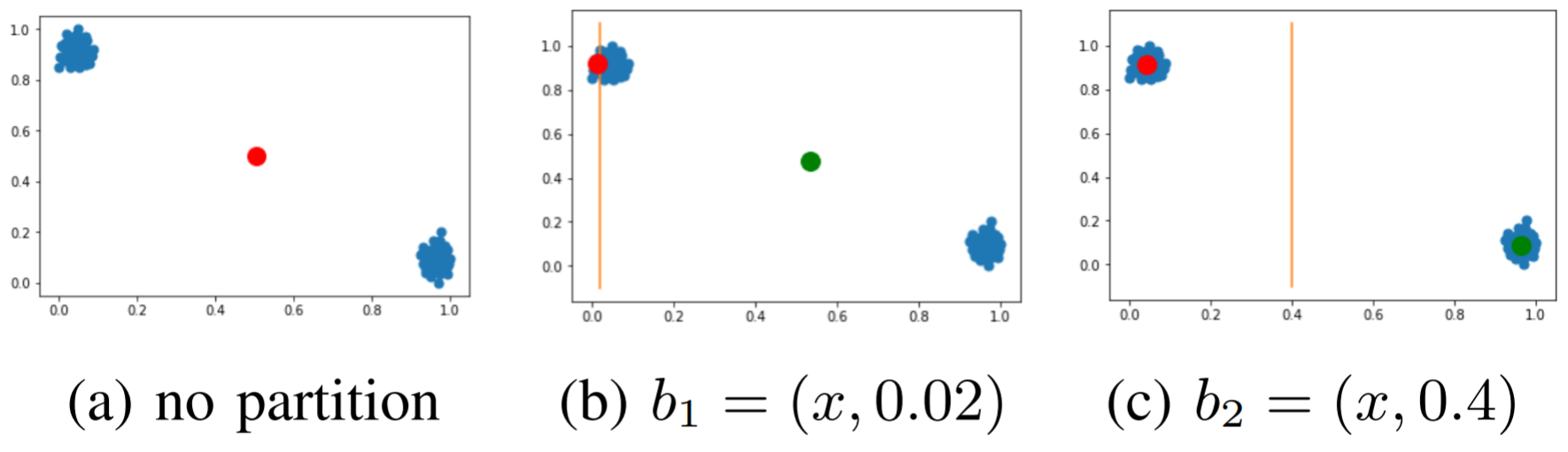}
    \caption{Examples of partitions. While $b_1$ makes the average
    distance to the left centroid small, the 
    average distance to the right centroid is large; $b_2$ is more balanced. }
    \label{fig:centroids}
\end{figure}

The basic outline of the algorithm follows the well known ID3 algorithm \cite{quinlan1986induction}: 
at each iteration, we choose the (locally) optimal feature/value and recursively partition the data until
some stopping criteria. For this, we must first quantify a `good' split. 
Given a set of points $D$, let $c_D$ be its centroid and $\delta_D$ be the average
distance between $c_D$ and all points in $D$. Define a boundary $b$ to be a pair
$(f,v)$, where $f$ is a feature and $v$ its value, and the left partition induced
by $b$ on $D$ as:
$$ D_{(b,l)} = \{ p | p\in D, p.f < v \}. $$
The right partition is $D_{(b,r)} = D \setminus D_{(b,l)}$. A boundary $b$ is a
good split if the partitions induced have smaller 
expected average distances to their respective
centroids (see Figure \ref{fig:centroids}).

Formally, we define the score of splitting at boundary $b$ as:
$$ S(D,b) = p_l (\delta_D - \delta_{D_{(b,l)}}) + p_r (\delta_D - \delta_{D_{(b,r)}}), $$
where $p_* = |D_{(b,*)} | / |D|$. Since $p_l + p_r = 1$, we can simplify the above as
$S(D,b) = \delta_D - (p_l \delta_{D_{(b,l)}} + p_r \delta_{D_{(b,r)}}))$. Finally, we define 
the reward as the score discounted by the (scaled) cost of obtaining that feature:
$$R(D,b) = (1-\alpha c(b.f))S(D,b). $$

At each node $\node$, we enumerate the features and evenly spaced
values from the domain of each feature. We then find the feature/value pair $b$ that maximizes $R(D_N,b)$. We stop when the number
of points is less than $\tau$, although other stopping criteria are possible. In terms of the complexity, 
we examine $|D_\node|$ elements at most $|\features| \cdot \ell $ times, where $\ell$ is the number of evenly spaced values.
For a tree of height $h$, this gives an upper bound of $O(|D||\features| \ell h)$ for construction.

\begin{algorithm}[t]
   \caption{build\_CBCT($D$)}
	Input: $D$ - dataset\\
	Output: $T$ - resulting tree 
	\begin{enumerate}
		\item if $|D| \le \tau$: return $T = (D,-,-)$
		\item for $f \in \features \setminus \bar\features$:
		\begin{enumerate}
			\item $\ell = \text{linspace}(f_{min}, f_{max})$
			\item for $v \in \ell$:
			\begin{enumerate}
				\item $b_{(f,v)} := (f,v)$
				\item compute $R(D, b_{(f,v)})$
			\end{enumerate}
		\end{enumerate}
		\item $b_{max} = \text{arg max}_b S(D,b)$
		\item set $c(b_{max}.f) := 0$
		\item $T_{l} := \text{build\_CBCT}(D_{(b_{max}, l)})$
		\item $T_{r} := \text{build\_CBCT}(D_{(b_{max}, r)})$
	\end{enumerate}
\end{algorithm}

\begin{figure}[t]
    \centering
    \subcaptionbox[0.3\linewidth]{points and boundaries}
    {\includegraphics[scale=0.22]{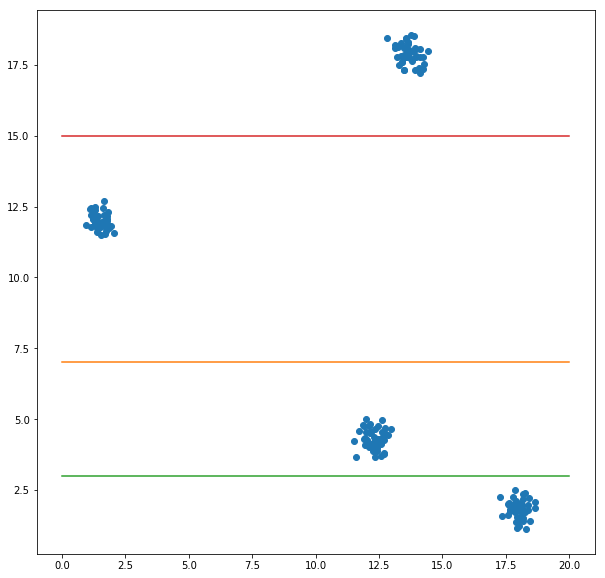}}
    
    \subcaptionbox[0.3\linewidth]{translated tree}
    {\includegraphics[scale=0.35]{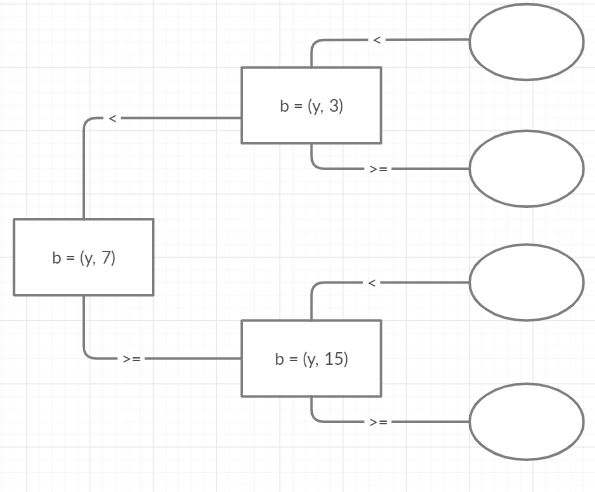}}
    
    \caption{Example of a CBCTree produced by a simple data set at uniform cost
    for each feature $x,y$. The rectangles
    denote decision nodes while the ovals are terminal nodes. Note the order
    of the splits; the first split divides the data in half, rather than singling out one cluster.
    Also, the $x$-values are never used as a boundary; while there are conflicts in the $x$-value between clusters,
    each subspace created does not have any conflicts.
    }
    \label{fig:my_label}
\end{figure}

\subsection{Suggested action and cluster prediction}
Given a set of $\bar\features$ strict subset of $\features$, we want to use a CBCTree $T$, as constructed
by the algorithm build\_CBCT, to suggest the next action to take. Since the features that decreases
the average distance the most are closer to the root of the tree, we can traverse down the tree
using features in $\bar\features$ and stop at the first node using a feature $f \notin \bar\features$
as the boundary.  Define $\nodes_{\bar\features,f,p}$ as the set of nodes reachable
with features $\bar\features \cup \{f\}$ following these rules of traversal at each node: 
1) if the splitting feature is known, go down the appropriate branch; 
2) if the splitting feature is $f$, go down both branches;
3) otherwise, add the node to $\nodes_{\bar\features,f,p}$. 
For each node $\node \in \nodes_{\bar\features,f,p}$, the \emph{similarity score}
of point $p$ with features $\bar\features$ is:
$$ \mathfrak{S}(\node, p, \bar\features) = \left( \frac{|D_{\node}|}{|D_{\text{total}}|} 
\sum_{p' \in D_{\node} } 
\sqrt{\sum_{f_i \in \bar\features } (p.f_i - p'.f_i)^2 } \right)^{-1},$$
where $|D_{\text{total}}| = \sum_{\node \in  \nodes_{\bar\features,f,p}} |D_{\node}|$ is the
number of total data points in $\node_{\bar\features,f,p}$.
The value $\mathfrak{S}$ is the reciprocal of the average L2 distance between each point of $D_N$ and $p$, only
using features $\bar\features$. Let $C_\node$ be the cluster defined by node $\node$ (that is, the region
defined by the boundaries to reach $N$), then $\mathfrak{S}(\node,p,\bar\features)$ is a measure of
how well $p$ `fits' into this cluster. Approximating the probability of $p$ was chosen from $C_\node$
as $\frac{|D_\node|}{|D_\text{total}|}$,
$\sum_{\node \in \nodes_{\bar\features,f}} \mathfrak{S}(\node,p,\bar\features)$ 
is the expected similarity score for revealing $f$ given we have $\bar\features$. The
higher this score, the more confident the model is that revealing $f$ is useful.
Similarly, let $\nodes_{\bar\features,p}$ be the set of nodes reachable by the same rules above, but
splitting on every feature $f \in (\features \setminus \bar\features)$. This represents the set of
all clusters that $p$ could belong to. For each $\node \in \nodes_{\bar\features,p}$, we compute
$\mathfrak{S}(N,p,\bar\features)$ and classify $p$ into $C_N$ that maximizes this value
(see algorithms \ref{algorithm:nf} and \ref{algorithm:n}).

\begin{algorithm}[ht]
\caption{compute\_$\nodes_{\bar\features,f}(T, p)$}
Input: $T$ - tree; $p$ - point \\
Output: $\nodes_f$
\label{algorithm:nf}
\begin{enumerate}
	\item if $T$ is empty: return $\{ \}$
	\item if $T$ is a leaf: return $\{ T.\text{node} \}$
	\item set $b$ to be the boundary at current node
	\item if $p[b.f]$ is known:
	\begin{enumerate}
		\item if $p[b.f] > b.v$: 
		return compute\_$\nodes_{\bar\features,f}(T.\text{right}, p)$
		\item else: return compute\_$\nodes_{\bar\features,f}(T.\text{left}, p)$
	\end{enumerate}
	\item else if $b.f = f$:
	\begin{enumerate}
		\item $\nodes_{\bar\features,f}^{(r)} = \text{compute\_}\nodes_{\bar\features,f}(T.\text{right},p)$
		\item $\nodes_{\bar\features,f}^{(l)} = \text{compute\_}\nodes_{\bar\features,f}(T.\text{left},p)$
		\item return $\nodes_{\bar\features,f}^{(r)} \cup \nodes_{\bar\features,f}^{(l)}$
	\end{enumerate}
	\item else: return $\{ T.\text{node} \}$
\end{enumerate}
\end{algorithm}

\begin{algorithm}[ht]
\caption{compute\_$\mathcal{N_{\bar\features}}(T,p)$}
Input: $T$ - tree; t - tuple \\
Output: $\mathcal{N}$
\label{algorithm:n}
\begin{enumerate}
	\item if $T$ is empty: return $\{ \}$
	\item if $T$ is a leaf: return $\{ T.\text{node} \}$
	\item let $b$ be the boundary at current node
	\item if $t[b.f]$ is known:
	\begin{enumerate}
		\item if $t[b.f] > b.v$: return compute\_$\mathcal{N_{\bar\features}}(T.\text{right},p)$
		\item else: return compute\_$\mathcal{N_{\bar\features}}(T.\text{left},p)$
	\end{enumerate}
	\item else:
	\begin{enumerate}
		\item $\mathcal{N}^{(r)} =$ compute\_$\mathcal{N_{\bar\features}}(T.\text{right},p)$
		\item $\mathcal{N}^{(l)} =$ compute\_$\mathcal{N_{\bar\features}}(T.\text{left},p)$
		\item return $\mathcal{N_{\bar\features}}^{(r)} \cup \mathcal{N_{\bar\features}}^{(l)}$
	\end{enumerate}
\end{enumerate}
\end{algorithm}

\section{Policy learning}
To solve MDP problems we estimate the optimal values of taking an action $a$ on state $s$. Q values are shaped by the expected future sum of rewards when taking $a$ on $s$ and then following the optimal policy thereafter. An optimal policy can be derived by greedily selecting the actions with the highest values at each state which satisfies the Bellman equation:
$$Q^*(s,a)=\EX_{s'} \left[ R + \gamma \max_{a'} Q^*(s', a' | s, a) \right]$$

We train a Deep Q Network (DQN) to schedule features that maximizes the ranking of the clusters while lowering cost incurred by the agent.

\subsection{Deep Q-learning}
Deep Q-learning introduces a target network, with parameters $\psi$, which follows an online network, with parameters $\theta$ \cite{mnih2015control}. The target network copies the parameters $\theta$ of the online network every $\tau$ steps, and for every other step, $\psi$ is fixed. The target used by the DQN is:
$$Y^Q_t = R_{t+1} + \gamma \max_{a \in \actions} Q_{\psi}(s_{t+1}, a)$$
where $\gamma$ is a scalar step size and $\actions$ is the set of available actions to take at state $s$. To optimize the agent, we minimize the Bellman mean squared error $\ell$ for a batch of transitions:
$$\ell_\theta = \EX \left[\left(Y_t^Q - Q_{\theta}(s_t, a_t) \right)^2 \right]$$

\subsection{Double Deep Q Networks}
In Q-learning and DQN, using the $\max$ operator to select and evaluate actions leads to over-optimistic value estimates \cite{hasselt2010doubleqlearning}. Double DQN is a technique which reduces this bias by rewriting the Q-value equation with respect to both online network $Q_{\theta}$ and target network $Q_{\psi}$ \cite{hasselt2016doubledqn}. They define the target $Y^Q_t$ as:
$$Y^Q_t = R_{t+1} + \gamma Q_{\psi}(s_{t+1}, \argmax_{a \in \actions} Q_{\theta}(s_{t+1}, a))$$

\subsection{Dueling Deep Q Networks}
Dueling DQN stabilizes and accelerates training by decomposing the Q-value function into value and advantage functions \cite{wang2016duelingdqn}. The network outputs an estimate of the value stream $V_{\theta}$ and advantage stream $A_{\theta}$ to construct the Q-value function as:
$$Q_{\theta}(s,a)=V_{\theta}(s)+ \left(A_{\theta}(s,a)-\frac{\sum_{a'}A_{\theta}(s,a')}{|\actions|}\right)$$

\paragraph{State}
In the DQN case, we follow a similar definition to $\states$ as defined in the Problem Formulation section. Since each state $s_{t,p}$ is composed of point $p$, learning Q-values over the states in $\states$ is a difficult task for this problem; small perturbations in the values of $p$ can change the order of the rankings. To reduce the state space of $\states$, for each state $s_{t,p} = (p,\bar\features_p, \bar c_p)$, we substitute $p$ with a multi-hot encoding $e_t$ to denote the features which are known at time step $t$.

\paragraph{Action selection}
Recall that each action in $\actions_f$ is a request for feature and the episode terminates when the agent chooses the action $\actions_t$. To restrict the DQN to perform only actions in $\actions_f$, we define the mask $m_t$ to be $m_t=(\mathbf{1} - e_t)$ where $\mathbf{1}$ is a vector of all ones. Additionally, $1$ is appended to $m_t$ to resemble the action $\actions_t$. We redefine the target function to include only available actions $\actions$ by applying the mask:
$$Y^{Q_\actions}_t = Y^Q_t \odot m_t$$

\begin{algorithm}[hb]
\caption{step(s,a)}
Input: $s = (p, \bar\features, \bar c)$ - 
state; $a \in \actions$ - action \\
Output: $s'$ - next state; $r$ reward
\begin{enumerate}
    \item if $a = \actions_t$:
    \begin{enumerate}
        \item $r = - S(\bar\features,p)$
    \end{enumerate}
    \item else:
    \begin{enumerate}
        \item $r = -\alpha c(a)$
        \item $s' = (p, \bar\features \cup \{ a \}, \bar c + c(a) )$
    \end{enumerate}
    
\end{enumerate}
\label{algorithm:step}

\end{algorithm}

\begin{figure}[ht]
    \centering
    \includegraphics[scale=0.23]{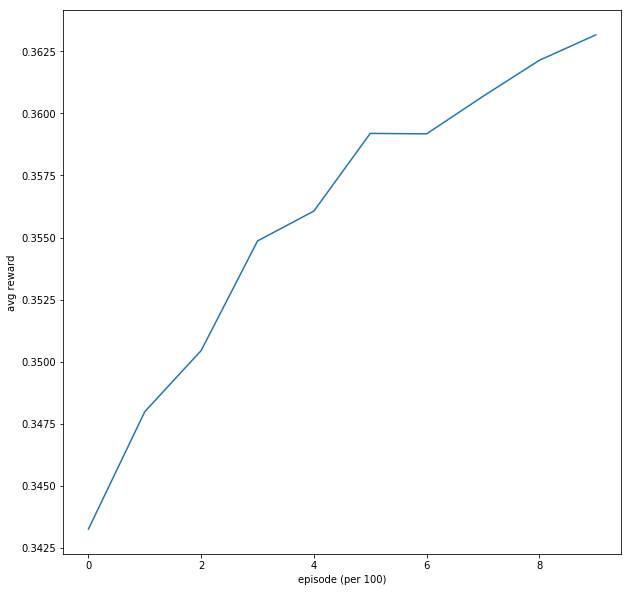}
    \caption{Plot of the average reward per 100 episodes during training for DQN using
    the data set heartfail. }
    \label{fig:RL}
\end{figure}

\paragraph{Reward}
To calculate the reward $r(s, a)$ received upon performing $\actions_t$ (See Algorithm \ref{algorithm:step}), we compute $S(\bar\features, p)$ as defined in Equation \ref{equation:score}. Concretely, the score function in Equation \ref{equation:score} is dependent on the implementation of the true ranking $\rank_{\features, i}(p)$ and predicted ranking $\rank_{\bar\features, i}(p)$ functions.
Let $$ d(p,p',\bar\features) = \sqrt{\sum_{f\in \bar\features} (p'.f - p.f)^2} $$

be the distance between $p$ and $p'$, using only $\bar \features$. To derive the true ranking $\rank_{\features, i}(p)$, we first cluster $D$ via K-means, producing cluster centroids $C_1,\dots,C_m$. We then order each
cluster by $d(p, C_i, \features)$, defining this order as the true rank $\rank_{\features,i}$. Similarly, given some $\bar\features$,
we order each cluster by $d(p, C_i, \bar\features)$, defining this order as the predicted rank $\rank_{\bar\features,i}(p)$.
Other implementations of clustering/ranking is possible; for instance, we could use the Gaussian mixture model and rank
the clusters by likelihood of observing $p$ given the distribution of a cluster.

\subsection{Comparison to inductive learning}
While both the CBCTree and DQN model aims to solve Problem \ref{problem:main}, there are several differences between
the two models. First, the CBCTree both clusters the data and provides a schedule of
feature updates, whereas our DQN model requires the clusters to be provided externally.
Consequentially, the results of CBCTree are more interpretable as the boundaries of the clusters are
explicitly computed. On the other hand, the `black box' approach of the
DQN model is more flexible since the external clustering algorithm is exchangeable.
Second, the CBCTree learns a hypothesis that is point (datum) wise, while our DQN model uses
binary features. We discuss the experimental implications in the next section.

The two models also differ in behavior when the features revealed are not following the model's learned policy. The suggested next feature update is fixed for the CBCTree to be the first 
unknown feature encountered when traversing the tree. Any other revealed feature is
used to compute the similarity score $\mathfrak{S}$. In Q-learning,
the Q-value function is defined for all possible states; thus, the model can dynamically
suggest the next update regardless of whether the users follow the learned policy.

\section{Analysis}

\begin{figure*}[t]
    \centering
    \subcaptionbox[0.3\linewidth]{heartfail}
    {\includegraphics[scale=0.25]{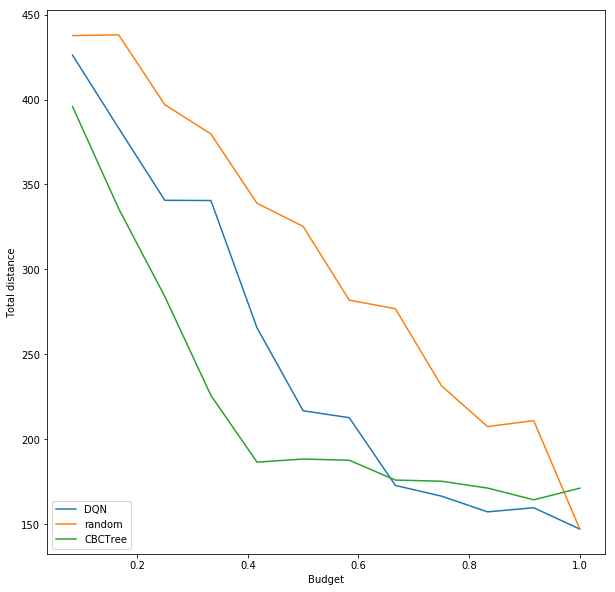}}
    \subcaptionbox[0.3\linewidth]{breastcancer}
    {\includegraphics[scale=0.25]{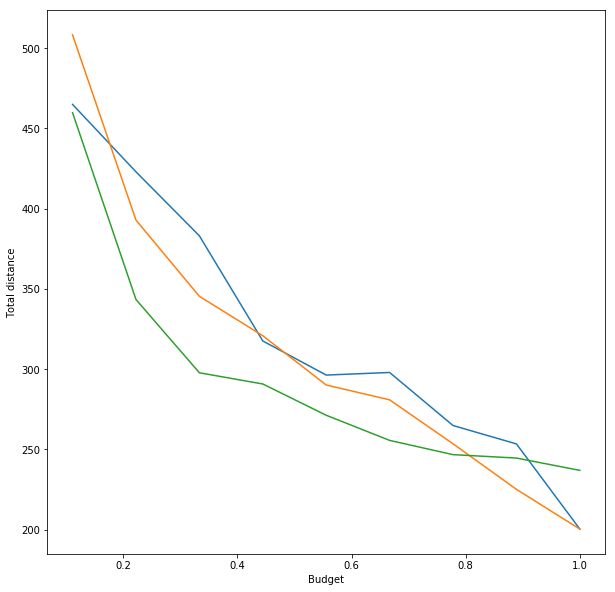}}
    \subcaptionbox[0.3\linewidth]{liver}
    {\includegraphics[scale=0.25]{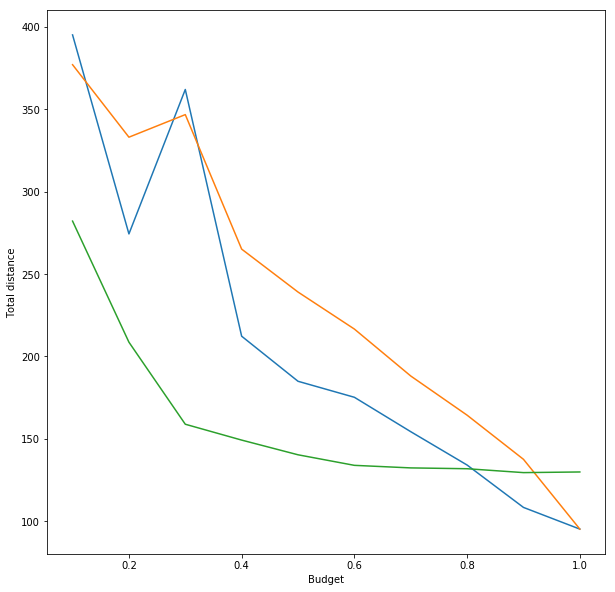}}
    
    \subcaptionbox[0.3\linewidth]{hcv}
    {\includegraphics[scale=0.25]{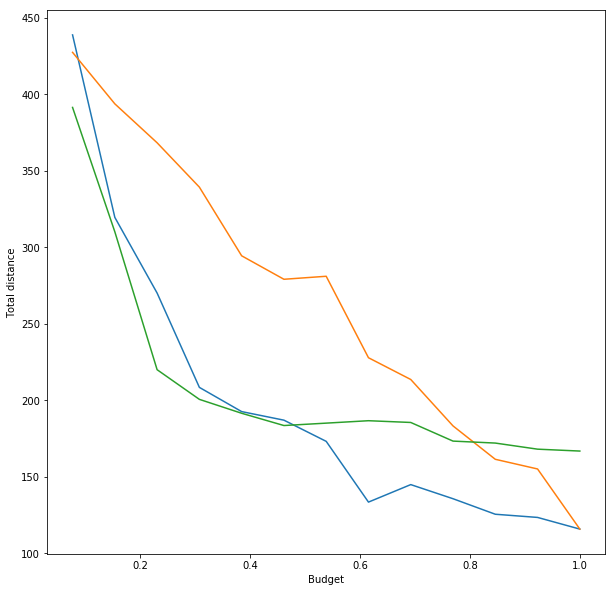}}
    \subcaptionbox[0.3\linewidth]{heart}
    {\includegraphics[scale=0.25]{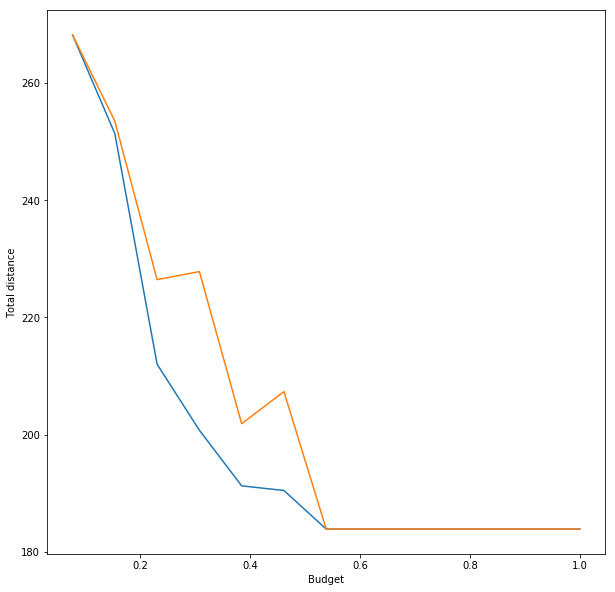}}
    \caption{Comparison of CBCTree, DQN, and random models}
    \label{fig:main_test}
\end{figure*}
We compare the performance of CBCTree, DQN, and an agent that takes random actions at
equal probability over all possible remaining actions at each step.
The first goal is to measure whether or not we can perform better than selecting
random features to reveal in real-life data sets. The second goal is to measure
which model performs the best; in particular, whether point-wise policies outperform
feature-wise policies.

\subsection{Environment Setup}
We use several publicly available data sets \cite{Dua:2019}, which are
summarized in Table \ref{tab:datasets}. For each set, we normalize 
data with its mean and range then divide the
data into training $D_{tr}$ and test $D_{t}$ sets. Other than `heart', all costs
are set uniformly to be $1/m$, where $m$ is the number of features. For heart, we
use the provided costs from Ontario Health Insurance Plan (OHIP)'s fee schedule. 
Furthermore, we used the provided `groups', which are sets of attributes such that
revealing one feature in a group will reveal the rest of the features in that group essentially
at no additional cost. For this, we omit the CBCTree as we have not yet implemented this feature
yet, although such scheduling is extremely natural for reinforcement learning.

\begin{table}[h]
    \centering
    \begin{tabular}{|c|c|c|c|}
        \hline
        Name & feat & \#train & \#test   \\
        \hline
        heart & 13 & 243 & 60 \\
        breastcancer & 9& 547 & 136 \\
        hcv & 13 & 493& 122 \\
        heartfail & 12& 240& 59 \\
        liver & 10& 464& 115\\
        \hline 
    \end{tabular}
    \caption{Summary of data sets}
    \label{tab:datasets}
\end{table}

We use the training set to construct each model at some fixed budget.
While some hyper-parameter tuning was performed to ensure that
the models converged, we did not do extensive tuning. The same hyper-parameters
are used for each data set. We also found that smaller DQN networks performed similarly to larger networks. 
For each instance $p$ in the test set, we start at the initial state
$(p,\phi,0)$ and apply the learned policies, providing the feature value as needed.
For the DQN and random agent, we first cluster the training set into $K$ clusters
using scikit-learn \cite{sklearn_api}. 

Recall that $$d(p,p',\bar\features) = \sqrt{\sum_{f\in \bar\features_p} (p'.f - p.f)^2}$$ 
is the distance between $p$ and $p'$ only using the revealed features $\bar\features$.
For the DQN and random models, after applying the learned policies, 
we predict $k=5$ closest points in $D_{tr}$ for each $p \in D_{t}$ 
by finding $k$ points $p_i \in D_{tr}$ such that $p_i$ has the $i$-th minimal value
$d(p,p_i,\bar\features)$.
Note that when $\bar\features_p = \features$, this is the $k$ closest points
in the training set to $p$. For the CBCTree, we still predict $k$ closest points,
but only within the predicted cluster. This is to better preserve the hard boundaries
that the CBCTree computes, since cluster dissimilarity is less clear on the CBCTree. 
Finally, we report $\sum_i d(p,p_i, \features)$, the sum of the true distances
between $p$ and $p_i$'s.

\subsection{Discussion}
Figure \ref{fig:main_test} shows the results of the three models on the data sets. For each plot,
the x-axis shows the allotted budget and the y-axis shows the sum of the $k$ true distances described
in the previous section; thus, lower points on the plot are better. 

There are several interesting observations we can make from these experiments. Firstly, with the exception
of breastcancer, the CBCTree and DQN models outperform the random agent. This is an important result because
it confirms that similar or `close' points can be identified without requiring the whole feature set, and additionally holds when the importance of features are non-uniform. This is already a well-established concept in classification
(for example, feature selection), but it is not immediately clear that such techniques are applicable when determining
similarity. Intuitively, if two entities are known to be similar in some aspect (eg. medical condition), then they should also be similar
in relevant features to that aspect, even if other unrelated measures differ (eg. hair colour); our experiments give evidence
that this intuition holds.

\begin{figure}[tb]
    \centering
    \includegraphics[scale=0.23]{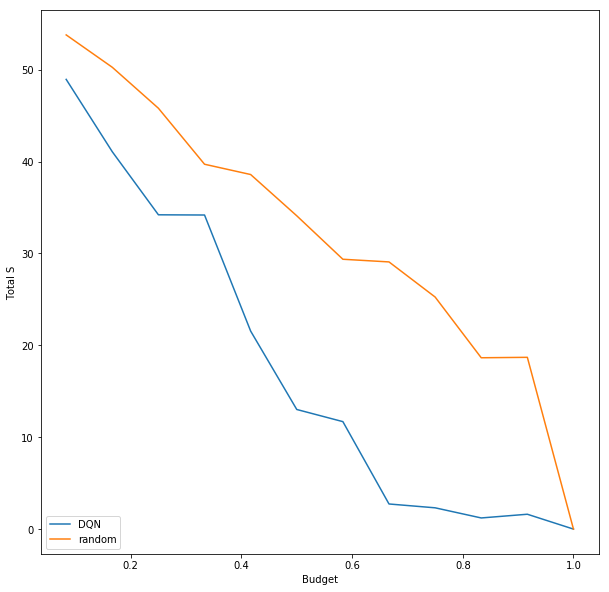}
    \caption{$\sum_{p\in D_{t}} S(\bar\features, p)$ for heartfail}
    \label{fig:Stotal}
\end{figure}

Secondly, the fact that we are able to retrieve nearby points by optimizing
Equation \ref{equation:score} shows that ranking the clusters is a useful proxy to solve Problem \ref{problem:main}.
Figure \ref{fig:Stotal} shows this in detail by plotting $\sum_{p\in D_{t}} S(\bar\features, p)$ for DQN and random agents 
for the data set heartfail. The shapes of both graphs are remarkably similar, which would be the case if they were
indeed closely related.

Thirdly, we can see that CBCTrees decrease distances faster at lower costs, but perform worse at higher costs. This is
due to the soft/hard boundaries discussed above. In particular, if the data has many points whose membership to a single
cluster is ambiguous, then hard boundaries may cause close points to be classified into different clusters. The difference
in the type of hypothesis learned could also explain why the CBCTrees perform better at lower costs. Since the DQN model uses 
binary features and we are always starting from an empty initial state, we are essentially using the Q-value learned by the
agent to select features. At lower budgets, selecting features globally is much less flexible than the point-wise method of the
CBCTree, but the difference is less pronounced as we increase the budget. One notable case where the DQN model perform even
worse than the random agent is breastcancer. The fact the CBCTree performs the best could imply that there exists data sets
whose features are equally important globally, but not locally conditioned on being in some subspace.

\section{Future works}
In the future, there are several extensions we plan to pursue. Firstly, from an algorithmic perspective, solving Problem
\ref{problem:main} allows for us to search for the K nearest neighbours (KNN) when there are missing features. One application
of this is to extend the current algorithm to work over time series data (possibly by applying a heavier cost to entries further
back in the past as they may be harder to retrieve than more current entries) and apply the KNN classifier algorithm along with
dynamic time warping, which has been shown to outperform many more sophisticated models \cite{dtw}.

A second direction is to consider how such human-in-the-loop feature updates can interact with completely automatic feature
completion models, such as low rank matrix completion. Since we are identifying nearby points, this could, in theory, help
algorithms determine some range of possible values. One experiment would be, for instance, to see if following the schedule
of updates provided by either the CBCTree or DQN models can improve the performance of a completion algorithm compared to
randomly choosing values to supply.

\section{Conclusion}
In this paper, we have proposed test-cost sensitive methods to identify similar or `close' points, given a new point with unknown feature values. We introduced two domain-independent methods that utilize our proposed MDP framework to optimize total feature cost and nearby point similarity. In five real-world datasets, both DQN and CBCTree outperform random policies in revealing features that maximize nearby point similarity. Furthermore, we found that Deep Q-learning performs well with only a few parameters in our MDP framework, enabling deployment in real-world settings. 

Although there is a growing body of research in test-cost sensitive methods, current efforts are focused on the context of classification. We hope that test-cost sensitive methods for identifying nearby points introduce a motivated and challenging environment for future research.

\bibliography{biblio}

\end{document}


\newpage
\section{Technical Appendix}
\label{sec:supplemental}

\section{Hyper-parameters for DQN}

\begin{table}[h]
\begin{tabular}{ |r|c| } 
 \hline
 Parameter & Value \\
 \hline
 \hline
 Episodes & 4000 \\
 \hline
 Hidden sizes & 1st layer: 128, 2nd layer: 256\\
 \hline
 Learning rate & 0.01 \\
 \hline
 Discount rate $\gamma$ & 0.8 \\
 \hline
 Initial $\epsilon$ & 1.0 \\
 \hline
 $\epsilon$ decay rate & 0.999 \\
 \hline
\end{tabular}
\caption{Parameters used in the experiment section for DQN.}
\end{table}
\subsection{Parameter selection}
For $\gamma$, we've tried [0, 0.3, 0.5, 0.8, 1]. 
We've assigned $\epsilon$ decay based on the number of epochs. In this case, we chose the decay to be $0.999$ since as the number of episodes approach $4000$, $\epsilon$ will approach $0.999^{4000} \cdot 1.0 = 0.018$.

We have also tried a larger network with hidden sizes 128, 258, 512, 1024, and 2048. We trained the network with 24,000 epochs, however, we did not find a significant difference in performance compared to the network with only two hidden layers.

\section{Hyper-parameters for CBCTree}
\begin{table}[h]
\begin{tabular}{ |r|c| } 
 \hline
 Parameter & Value \\
 \hline
 \hline
 Min size clusters $\tau$ & 10 \\
 \hline
 cost scaler $\alpha$ & 1 \\
 \hline
\end{tabular}
\caption{Parameters used in the experiment section for CBCTree.}
\end{table}

\section{Dataset}
\subsection{Train/Validation Split}
We randomly split the dataset into 80\%/20\% for the training and validation splits, respectively. The validation set was used to tune the parameters, based on rewards received from the MDP.

\subsection{Computing Infrastructure}
We used a AMD Ryzen 7 1700 CPU with 32 GB of RAM and a Nvidia GTX 1080 GPU. The average training time with our setup was 15 minutes per 4000 episodes.